%% file: 00_IEEE_main.tex
\documentclass[conference]{IEEEtran}
\IEEEoverridecommandlockouts
\usepackage{cite}
\usepackage{amsmath,amssymb,amsfonts}
\usepackage{algorithmic}
\usepackage{graphicx}
\usepackage{textcomp}
\usepackage{xcolor}
\def\BibTeX{{\rm B\kern-.05em{\sc i\kern-.025em b}\kern-.08em
    T\kern-.1667em\lower.7ex\hbox{E}\kern-.125emX}}
\input{header}

\begin{document}


\title{ABN: Anti-Blur Neural Networks for Multi-Stage Deformable Image Registration}

\author{\IEEEauthorblockN{Yao Su} 
\IEEEauthorblockA{\textit{Worcester Polytechnic Institute} \\
Worcester, MA, USA \\
ysu6@wpi.edu
\and
\IEEEauthorblockN{Xin Dai} 
\IEEEauthorblockA{\textit{Worcester Polytechnic Institute} \\
Worcester, MA, USA \\
xdai5@wpi.edu
\and
\IEEEauthorblockN{Lifang He} 
\IEEEauthorblockA{\textit{Lehigh University} \\
Bethlehem, PA, USA \\
lih319@lehigh.edu
\and
\IEEEauthorblockN{Xiangnan Kong} 
\IEEEauthorblockA{\textit{Worcester Polytechnic Institute} \\
Worcester, MA, USA \\
xkong@wpi.edu
}}}}
}

\maketitle

\input{01_abstract_v3}

\input{02_intro_v3}

\input{03_problem_def3}

\input{04_method}

\input{05_experiment_appendix_merged}

\input{06_related_work}

\input{07_conclusion}

\input{10_ack}

\bibliographystyle{IEEEtran}
\bibliography{reference}
\input{fig_res_mind}
\end{document}

%% file: header.tex
\usepackage{amsthm}
\usepackage{amsmath}
\usepackage{bm}
\usepackage{color}
\usepackage{etex}
\usepackage{qtree}
\usepackage{graphicx}
\usepackage{multirow}
\usepackage{multicol}
\usepackage{subfigure}
\usepackage{url}
\usepackage{thmtools}
\usepackage{ctable}
\usepackage{tabularx}
\usepackage{balance}    
\usepackage{booktabs}   
\usepackage{graphics}   
\usepackage{url}        
\usepackage{pifont}     
\usepackage{xcolor}      
\usepackage{xspace}
\usepackage{balance}
\newcommand{\ie}[0]{\textit{i.e.},\ }   
\newcommand{\eg}[0]{\textit{e.g.},\ }   

\usepackage[labelfont={bf,small},textfont={bf,small}]{caption}
\usepackage{pifont}
\newcommand{\cmark}{\ding{51}}%
\newcommand{\xmark}{\ding{55}}%

\newcommand{\eat}[1]{}



%% file: 01_abstract_v3.tex
\begin{abstract}
Deformable image registration, \ie the task of aligning multiple images into one coordinate system by non-linear transformation, serves as an essential preprocessing step for neuroimaging data. 
Recent research on deformable image registration is mainly focused on improving the registration accuracy using multi-stage alignment methods, where the source image is repeatedly deformed in stages by a same neural network until it is well-aligned with the target image. Conventional methods for multi-stage registration can often blur the source image as the pixel/voxel values are repeatedly interpolated from the image generated by the previous stage. However, maintaining image quality such as sharpness during image registration is crucial to medical data analysis.
In this paper, we study the problem of anti-blur deformable image registration and propose a novel solution, called Anti-Blur Network (ABN), for multi-stage image registration.
Specifically, we use a pair of short-term registration and long-term memory networks to learn the nonlinear deformations at each stage, where the short-term registration network learns how to improve the registration accuracy incrementally and the long-term memory network combines all the previous deformations to allow an interpolation to perform on the raw image directly and preserve image sharpness.
Extensive experiments on both natural and medical image datasets demonstrated that ABN can accurately register images while preserving their sharpness.

\end{abstract}

%% file: 02_intro_v3.tex

\section{Introduction}
\label{sec:intro}
\textbf{Background.} Image registration is an essential task in medical image analysis with many applications, including anatomical and functional studies~\cite{bai2017unsupervised,wang2017structural}, multi-modality fusion~\cite{cai2018deep} and diagnostic assistance~\cite{sun2009mining}.
It represents the process of estimating the transformation between two images and align them into one coordinate system. 
In the same coordinate system, the interference of viewpoints, motion and imaging modalities can be eliminated, thus allowing accurate quantification of changes in the position, size, and shape of anatomy and function.
In the diagnosis of pneumonia, a patient’s serial chest CT scans need to be aligned together to counteract breathing movement. 
This crucial preprocessing step helps doctors accurately locate tumors and shadows in the lung, and perform medical diagnosis.
The morphology of human organs is diverse, which presents a unique challenge to precise image registration.
Conventional methods for image registration often rely on simple linear transformation to optimize the similarity between source and target images, such as rigid transformations and affine transformations~\cite{avants2009advanced,jenkinson2001global}, which render coarse results, as shown in Figure~\ref{fig:family 1}. 
To address this limitation, deformable image registration has attracted increasing attention in recent years~\cite{rueckert1999nonrigid, thirion1998image, bajcsy1989multiresolution, avants2008symmetric,chen2013large,wulff2015efficient}, which intends to learn a nonlinear transformation to boost registration performance.

\input{fig_problem}

\textbf{State-of-the-Art.} 
With the emergence of deep learning, convolutional neural networks (CNNs) have recently been introduced for deformable image registration \cite{balakrishnan2019voxelmorph,de2017end,li2017non,zhao2019recursive, de2019deep} due to their superior capability. Conventional CNN-based methods~\cite{balakrishnan2019voxelmorph,de2017end,li2017non} focus on estimating the deformation by one-step to derive a registered image, where the transformation is directly predicted by a single network, as shown in Figure~\ref{fig:family 2}. However, this design is difficult to achieve satisfactory registration results, especially when there are complex and large deformations between source and target images.
To address this limitation, multi-stage registration methods~\cite{zhao2019recursive, de2019deep} have been developed based on cascaded neural networks, which tends to improve the registration accuracy by multiple stages of transformation, as shown in Figure~\ref{fig:family 3}. 
However, these methods often lead to blurry registered images and loss of details as they perform transformation only on the image generated by the previous stage, involving repeated interpolation to the final output. 
Maintaining image quality such as sharpness during image registration is crucial in many medical studies. 
In anatomical pathology studies, abnormalities (\eg tumors and lesions) in blurred images will be more challenging to localize due to weak edge information, thus affecting the accuracy of subsequent diagnosis.
Likewise, the precision of functional pathology diagnosis (\eg Alzheimer's disease) also relies on the integrity of the image information. Blurred images often lose their high-frequency components, making it difficult to measure variations in functional signals.

\textbf{Problem Definition.} In this paper, we study the problem of anti-blur deformable image registration, as shown in Figure~\ref{fig:1}. 
Our goal is to accurately align the source image with a target image, while maintaining the sharpness of the registered image in the process.

\textbf{Challenges.} Despite its value and significance, the anti-blur deformable image registration problem is very challenging due to its unique characteristics listed below: 
\begin{itemize}
    \item \textit{Transform Estimation:}
    Most conventional deformable image registration methods~\cite{balakrishnan2018unsupervised,balakrishnan2019voxelmorph,de2017end,li2017non} accomplish this task by directly predicting the transformation between source and target images. However, estimating transformation with one-step prediction can limit the registration accuracy, especially when there are complex and large deformations between images.
    Moreover, when processing high-dimensional images (\eg 3D MRI and CT), one-step transformation approaches require manipulating larger voxel quantities, producing lower accuracy and having limited capabilities.
    
    \item \textit{Repeated interpolations:} 
    Recent state-of-the-art methods~\cite{zhao2019recursive, de2019deep} use multi-stage design, which improves the registration accuracy to a certain extent. However, repeatedly transforming the image generated by the previous stage involves multiple interpolations of pixel/voxel value, which can result in a loss of image information. Specifically, the high-frequency components of the image will be smooth by multiple interpolations, leading to a blurred image. 
    
    \item \textit{Error Accumulation:}
    Conventional multi-stage registration methods~\cite{zhao2019recursive, de2019deep} are based on cascade design, which can cause errors to accumulate. An inaccurate deformation estimation from the previous stage can result in artifacts on the output image being propagated into the following stages, rendering an irreversible error. For example, if some regions of the input image are mistakenly moved out, the information contained in these regions cannot be recovered in the following stages.
\end{itemize}

In order to tackle these issues, we propose an Anti-Blur Network (ABN) for multi-stage deformable registration. Figure \ref{fig: family} illustrates the comparison between our approach and the state-of-the-art methods. 
Specifically, inspired by Long Short-Term Memory (LSTM), we introduce a pair of short-term registration and long-term memory networks to learn the nonlinear deformations at each stage, where the short-term registration network is used to capture the deformation between current warped (registered) and target images, and the long-term memory network combines all the previous deformations so that the transformations (involving interpolation) in subsequent stages consistently act on the source image, thus preserving the sharpness of the registered image to a greater extent.
Empirical studies on natural and medical image registration tasks demonstrated that ABN outperform both single-stage and multi-stage existing methods in terms of image registration accuracy and sharpness.
The main contributions are summarized as follows:

\begin{itemize}
    \item To the best of our knowledge, this is the first work to study 
    anti-blurring problem in multi-stage deformable image registration, which allows nonlinear transformations to be consistently performed on the source image in a multi-stage setting.
    \item We identify a new criterion to evaluate the performance of image registration by measuring the sharpness of registered image from information completeness viewpoint.
    \item Extensive experiments are conducted on the 2D face image registration task and 3D brain MRI registration task, and the results indicate that our proposed method significantly outperforms state-of-the-art alternatives in terms of both registration accuracy and image sharpness.
\end{itemize}
\input{fig_family}

%% file: fig_problem.tex

\begin{figure}[t]
  \centering
  \includegraphics[width=\linewidth]{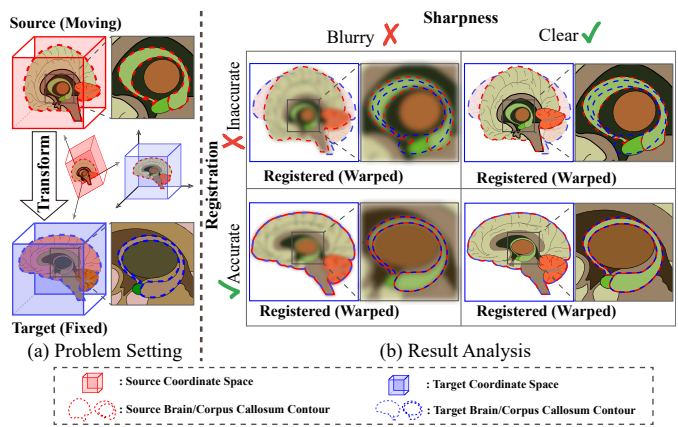}
  \vspace{-10pt}
  \caption{
  The problem of anti-blur deformable image registration.
  (a) Given a patient's brain MRI scan (the source image) and a template image of the standard brain (the target image), the goal is to transform the source image non-linearly so that it is aligned with the target image and the sharpness of the registered image should be preserved in the process. Example of different possible results are shown in (b). The bottom-right box is the ideal result: the registered image should be well-aligned with the target image while preserving the sharpness.
  }
  \label{fig:1}
  \vspace{-15pt}
\end{figure}

%% file: fig_family.tex


\begin{figure}[t]
    \centering
    \subfigure[Single-stage affine (linear) registration
    \cite{avants2009advanced,jenkinson2001global}]{
        \includegraphics[width=0.95\linewidth]{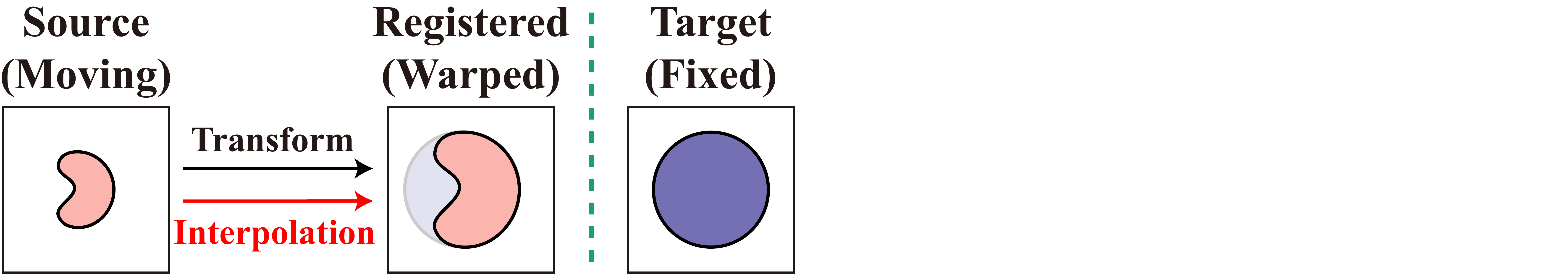}
        \label{fig:family 1}
    }
    
    \subfigure[Single-stage deformable (nonlinear) registration
    \cite{balakrishnan2019voxelmorph,de2017end,li2017non}]{
        \includegraphics[width=0.95\linewidth]{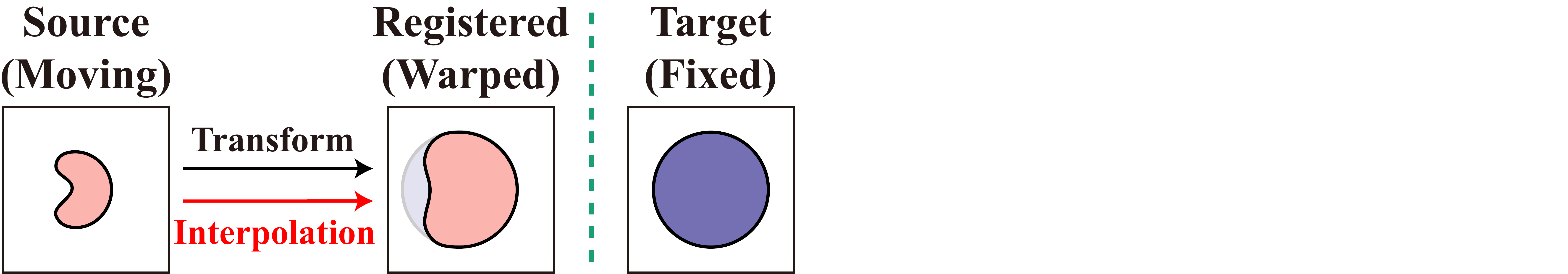}
        \label{fig:family 2}
    }
    \subfigure[Multi-stage deformable registration \cite{zhao2019recursive,de2019deep}]{
        \includegraphics[width=0.95\linewidth]{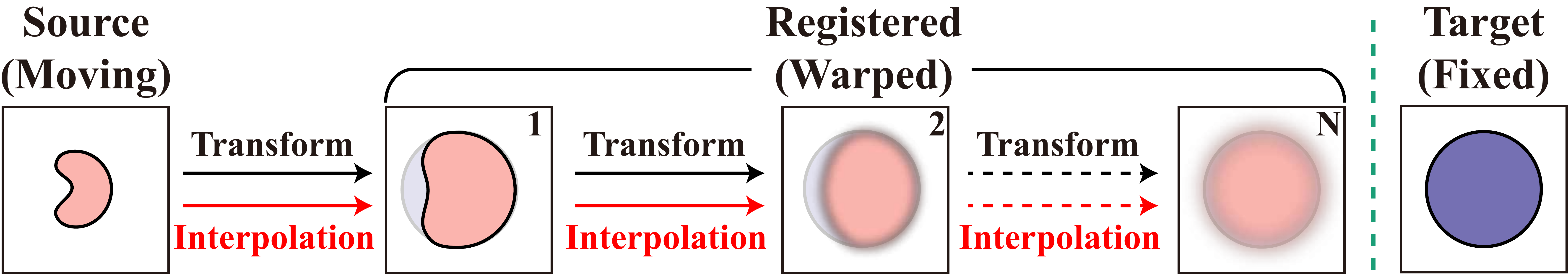}
        \label{fig:family 3}
    }
    \subfigure[Multi-stage anti-blur deformable registration (our method)]{
        \includegraphics[width=0.98\linewidth]{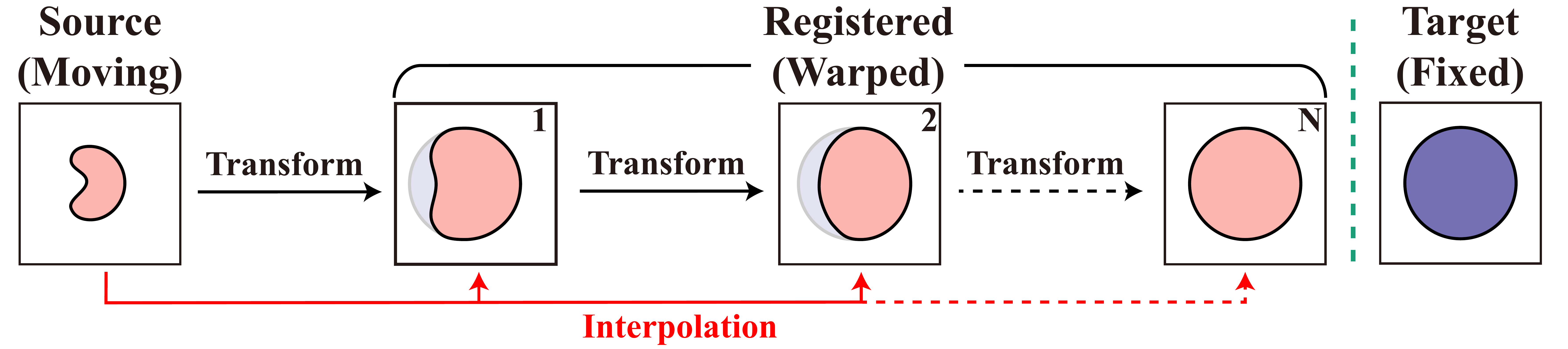}
        \label{fig:family 4}
        
    }
    \vspace{-10pt}
    \caption{Comparing different solutions to the deformable image registration problem.}
    \label{fig: family}
    \vspace{-15pt}
\end{figure}

%% file: 03_problem_def3.tex
\section{PRELIMINARIES} 
In this section, we first introduce related concept and notations, then define the anti-blur deformable image registration problem formally.

\subsection{Notations and Definitions}

In this paper, we mainly consider two scenarios: the 2D face registration and 3D brain MRI registration. In the following discussion, we introduce the concept of deformable registration using 3D brain MRI images, and it can be easily generalized to 2D image registration scenarios.

\noindent \textbf{Definition 1 (Source and target images).} Suppose we are given a training dataset $\mathcal{D} = \left\{\left(\mathbf{S}_i,\mathbf{T}_i\right)\right\}_{i=1}^{Z}$ that contains $Z$ pairs of training samples. Each pair contains a source image $\mathbf{S}_{i} \in \mathbb{R}^{W \times H \times D}$ (\eg the MRI scan of a patient's brain) and a target $\mathbf{T}_{i} \in \mathbb{R}^{W \times H \times D}$ (\eg a standard template of the brain). Here $W$, $H$ and $D$ denote the width, height and depth dimensions of the 3D images. Without loss of generality, here we assume that the source and target images are resized to the same dimension, \ie $W \times H \times D$. Typically, in the dataset $\mathcal{D}$, the target images in different pairs can be different. 
For example, in the human face registration task~\cite{kowalski2017deep}, we need to align the distorted image (\ie source image) with the normal image (\ie target image) for each person, where different people will have different face images. In the brain MRI registration task, however, all pairs in $\mathcal{D}$ can share a same target image, which is another case of the dataset $\mathcal{D}$. For example, in brain network studies~\cite{sun2009mining}, the functional MRI images (\ie source images) of all patients need to be aligned with a same template image (\ie target image), \eg MNI 152. We study both cases of $\mathcal{D}$ in this paper, and the experimental results are shown in the Section~\ref{section:Experiments}.
In the following discussion, we omit the subscript $i$ of $\mathbf{S}_{i}$ and $\mathbf{T}_{i}$ for simplicity.

\noindent \textbf{Definition 2 (Deformation field and warped image).} Deformation field $\mathbf{\Phi} \in \mathbb{R}^{W \times H \times D \times 3}$ is a tensor used to parameterized the nonlinear (deformable) transformation between $\mathbf{S}$ and $\mathbf{T}$. It defines how each voxel in $\mathbf{S}$ is displaced (in the X-, Y-, and Z-directions) with respect to its original position in $\mathbf{T}$. 
The warped image $\mathbf{W} = \mathcal{T}\left(\mathbf{S},\mathbf{\Phi}\right)$ is generated by applying the deformable transformation on the source image $\mathbf{S}$, where $\mathcal{T}(\cdot, \cdot)$ is the warping operator.
The voxel value of $\mathbf{W}$ at the 3D coordinate $(x, y, z)$ can be conceptually calculated as:
\begin{equation}
\label{equ:voxel_value}
\mathbf{W}(x,y,z) = \mathbf{S}(x+ \Delta x,y+\Delta y,z+\Delta z),
\end{equation}
where 
\begin{equation}
\label{equ:delta}
\Delta x = \Phi(x,y,z, 1), \hspace{6pt} \Delta y = \Phi(x,y,z, 2), \hspace{6pt} \Delta z = \Phi(x,y,z, 3).
\end{equation}
Typically, $(\Delta x,\Delta y,\Delta z)$ is a vector of continuous values, so is the 3D coordinate $(x+\Delta x,y+\Delta y,z+\Delta z)$. As a result, the voxel value $\mathbf{W}(x,y,z)$ cannot be directly calculated by Eq~\eqref{equ:voxel_value}. Instead, the image sampling kernel, \eg trilinear sampling (linear interpolation in the 3D coordinate space), is resorted to estimate the voxel value. Figure \ref{fig:mapping} demonstrates an example of deformable mapping and interpolation.

\subsection{Problem Formulation}
In this work, we consider the deformable registration task as a learning problem on neural networks. This is often modeled by learning a function: $f_{\theta}: \mathbb{R}^{W \times H \times D}\times \mathbb{R}^{W \times H \times D} \rightarrow \mathbb{R}^{W \times H \times D \times 3}$. Specifically, the function $f_{\theta}(\cdot, \cdot)$ takes the source image $\mathbf{S}$ and the target image $\mathbf{T}$ to predict the deformation field $\hat{\mathbf{\Phi}}$, and the warped image is $\hat{\mathbf{W}} = \mathcal{T}(\mathbf{S},\hat{\mathbf{\Phi}})$.
The optimal parameters $\theta^{*}$ can be found by solving the following optimization problem:
\begin{equation}
\begin{split}
\label{equ:goal_training}
\theta^{*} &=\underset{\theta}{\arg \min } \hspace{-3pt} \sum_{\left(\mathbf{S},  \mathbf{T}\right)\in \mathcal{D}}\left[ \mathcal{L} \left( \hat{\mathbf{W}}, \mathbf{T}  \right)\right] \\
& = \underset{\theta}{\arg \min } \hspace{-3pt} \sum_{\left(\mathbf{S},  \mathbf{T}\right)\in \mathcal{D}}\Big[ \mathcal{L} \Big( \mathcal{T} \Big(\mathbf{S}, \hspace{0pt} f_{\theta}\left(\mathbf{S}, \mathbf{T}\right)\Big), \mathbf{T}  \Big)\Big],
\end{split}
\end{equation}
where the image pair $(\mathbf{S},\mathbf{T})$ is sampled from the training dataset $\mathcal{D}$, and $\mathcal{L}(\cdot, \cdot)$ is image dissimilarity criteria, \eg mean square error.

Recent studies \cite{zhao2019recursive,de2019deep} proposed to solve Problem (\ref{equ:goal_training}) in a multi-stage fashion to achieve high registration accuracy, which consist of $N$ cascaded networks $\{f_k\}_{k=1}^N$. 
In the $k$-th cascaded stage, $\mathbf{\Phi}^{k}=f_{k}\left(\mathbf{W}^{k-1}, \mathbf{T}\right)$, where $\mathbf{W}^{k-1}=\mathcal{T}\left(\mathbf{W}^{k-2}, \mathbf{\Phi}^{k-1}\right)$ is the intermediate warped image. 
However, the warping operator $\mathcal{T}$ involves an interpolation, \eg trilinear sampling; thus, the final result $\mathbf{W}^{N}$ can be blurred by $N$ warping operations (\ie $N$ times interpolation).

Our aim is to boost the performance of deformable image registration while preserving the sharpness of registered images to ensure their information integrity. Specifically, we seek a novel multi-stage solution $\mathbf{\Phi}^{k}=f_{k}\left(\mathbf{W}^{k-1} \otimes \mathbf{S}, \mathbf{T}\right)$ that allows to jointly use the source image $\mathbf{S}$ and the warped image $\mathbf{W}^{k-1}=\mathcal{T}\left(\mathbf{S},\mathbf{\Phi}^{k-1}\right)$ to minimize the sharpness loss during the successive warping processes. Notice that the warping operation always performs on the source image $\mathbf{S}$ instead of the previous warped images. Therefore, only one interpolation is needed to produce the final warped image $\mathbf{W}^{N}$, which preserves the sharpness of $\mathbf{W}^{N}$.

\input{fig_mapping}

%% file: fig_mapping.tex
\begin{figure}[t]
  \centering
  \includegraphics[width=\linewidth]{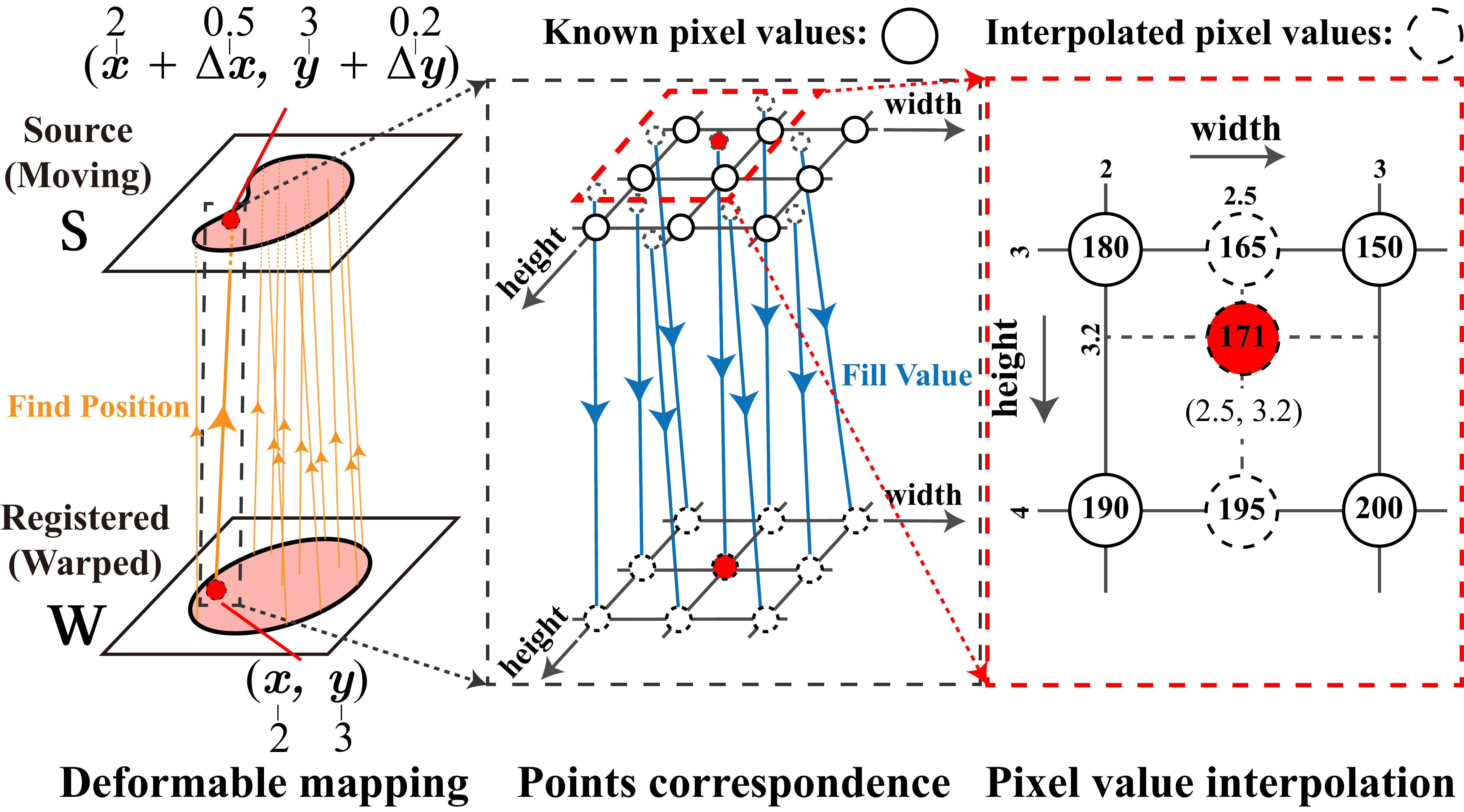}
  \vspace{-10pt}
  \caption{An example of deformable image registration in 2D space. On the left, the yellow lines denote the coordinate correspondences between the warped image and the source image: $(x,y)\mapsto(x+ \Delta x,y+\Delta y)$ . In the middle, the blue lines represent the pixel value of each grid point of the warped image filling by a corresponding interpolated point from the source image: $\mathbf{S}(x+ \Delta x,y+\Delta y)\mapsto\mathbf{W}(x,y)$. On the right, the pixel value of an interpolated point is estimated from its adjacent points. }
  \label{fig:mapping}
  \vspace{-10pt}
\end{figure}

%% file: 04_method.tex
\input{fig_network2}

\section{Our Approach}
\label{sec:method}
\noindent\textbf{Overview.} Figure~\ref{fig:network} presents an overview of the proposed ABN framework for anti-blur deformable image registration. Our method is a multi-stage deep neural network consisting of three main components: 1) \emph{Short-Term Registration Network} takes the previous warped image $\mathbf{W}^{k-1}$ and the target image $\mathbf{T}$ as inputs, and generates a current deformation field $\mathbf{\Phi}_\text{i}^{k}$; 2) \emph{Long-Term Memory Network} takes the previous combined deformation field $\mathbf{\Phi}_\text{c}^{k-1}$ and the current deformation field $\mathbf{\Phi}_\text{i}^{k}$ as input, and generates an updated combined deformation field $\mathbf{\Phi}_\text{c}^{k}$; and 3) \emph{Spatial Transformation Layer} performs the nonlinear transformation on the source image $\mathbf{S}$ using $\mathbf{\Phi}_\text{c}^{k}$ to produce the warped image $\mathbf{W}^{k}$. The whole framework is trained using backpropagation in an end-to-end and unsupervised fashion.
Next we introduce the details of each component and the training process.
\subsection{Short-Term Registration Network: $f_{S}$}
The short-term registration network $f_{S}(\cdot, \cdot)$ is designed to gradually deform the source image to maximize its similarity with the target image. At each stage, it predicts a current deformation field $\mathbf{\Phi}_\text{i}^{k}$ and only relies on the previous warped image $\mathbf{W}^{k-1}$ and the target image $\mathbf{T}$.
Decomposing a complex registration task into multiple simple short-term registration tasks can substantially reduce the burden on the network, allowing it to control the deformation process more locally and accurately and avoid under- or over-deformation of the image.
Specifically, we adopt U-Net \cite{ronneberger2015u} as the base network to learn $f_{S}(\cdot, \cdot)$, which is the state-of-the-art architecture widely used in image registration and semantic segmentation. 
$f_{S}(\cdot, \cdot)$ follows a shared-weight design, which means that $f_{S}(\cdot, \cdot)$ is repetitively applied across stages with the same parameters. It can be formalized as:
\begin{equation}
\mathbf{\Phi}_\text{i}^{k}=f_{S}\left(\mathbf{W}^{k-1}, \mathbf{T}\right),
\end{equation}
where $\mathbf{\Phi}_\text{i}^{k}$ is the outputted deformation field in the $k$-th stage for $k=1, \cdots, N$ and $\mathbf{W}^{0} = \mathbf{S}$.

\subsection{Long-Term Memory Network: $f_{L}$}
In each stage, after obtaining the current deformation field $\mathbf{\Phi}_\text{i}^{k}$ from $f_{S}(\cdot, \cdot)$, we are seeking a function to fuse all previous deformations that allows to jointly use the source image $\mathbf{S}$ and the previous warped image $\mathbf{W}^{k-1}$ to preserve the image sharpness for better details.
In other words, this function needs to recursively combines the current deformation with all previous deformations in order to learn the transformation mapping between $\mathbf{S}$ and $\mathbf{T}$ at each stage.
Therefore, the estimation of image pixel/voxel values (\ie trilinear sampling) will be allowed to be directly performed on $\mathbf{S}$ to avoid image sharpness loss caused by multiple interpolations.
To achieve this goal, we design a long-term memory network $f_{L}(\cdot, \cdot)$ to predict the combined deformation field $\mathbf{\Phi}_\text{c}^{k}$ as follows:
\begin{equation}
\mathbf{\Phi}_\text{c}^{k}=f_{L}\left(\Phi_{c}^{k-1}, \mathbf{\Phi}_\text{i}^{k}\right).
\end{equation}
When $k=1$, an initialized deformation field $\mathbf{\Phi}_\text{c}^{0}$ with zero displacements and the current deformation field $\mathbf{\Phi}_\text{i}^{1}$ are inputs of $f_{L}(\cdot, \cdot)$.
This layer serves as a bridge between the registration module and the spatial transformation module. 
Notice that $f_{L}(\cdot, \cdot)$ is the main component for anti-blur, otherwise, the spatial transformation can only act on $\mathbf{W}^{k-1}$.

Similar to $f_{S}(\cdot, \cdot)$, we adopt an U-Net based CNNs to learn $f_{L}(\cdot, \cdot)$ with a weight sharing across each stage. 


\subsection{Spatial Transformation Layer}
An essential step required for image registration is to reconstruct the warped image from the source image by the warping operator. 
Based on the combined deformation $\mathbf{\Phi}_\text{c}^{k}$, we introduce a spatial transformation layer that resamples pixels/voxels from the source image into the uniform grid to obtain a warped image through $\mathbf{W}^{k} = \mathcal{T}(\mathbf{S}, \mathbf{\Phi}_\text{c}^{k})$. Based on the definition of warping operator in Eq.~(\ref{equ:voxel_value}), we have
\begin{equation}
     \mathbf{W}^{k}(x, y, z) = \mathbf{S}(x+ \Delta x^{k},y+\Delta y^{k},z+\Delta z^{k}),
     \label{equ:voxel_value_k}
\end{equation}
where $\Delta x^{k} = \mathbf{\Phi}_\text{c}^{k}(x,y,z, 1),~\Delta y^{k} = \mathbf{\Phi}_\text{c}^{k}(x,y,z, 2)$, and $\Delta z^{k} = \mathbf{\Phi}_\text{c}^{k}(x,y,z, 3)$.


To enable the success of gradient propagation in this process, we employ a differentiable warping operator based on trilinear interpolation inspired by \cite{jaderberg2015spatial}. That is, 
\begin{equation}
\begin{split}
& \mathbf{W}^{k}(x ,y ,z) = \sum_{n}^{W} \sum_{m}^{H} \sum_{l}^{D} \mathbf{S}(n,m,l)
\max(0,1-|x+ \Delta x^{k}-n|) \\ 
& \max (0,1-|y+ \Delta y^{k}-m|)
\max (0,1-|z+ \Delta z^{k}-l|).
\end{split}
\end{equation}
Notice that Eq.~(\ref{equ:voxel_value_k}) always performs nonlinear warping on the original source image $\mathbf{S}$, instead of the warped image generated by the previous stage. Therefore, only one interpolation is needed to yield a warped image $\mathbf{W}^{N}$ to preserve its sharpness.

\subsection{Unsupervised End-to-End Training}
We train our ABN model in an unsupervised setting by minimizing the following objective function
\begin{equation}
\label{eq:loss}
\underset{\mathbf{\Phi}_\text{c}^{1}, \cdots, \mathbf{\Phi}_\text{c}^{N}}{\min} \mathcal{L}_{\operatorname{sim}} \left(\mathbf{W}^{N}, \mathbf{T}\right)+ \sum_{k=1}^{N}\lambda \mathcal{R}(\mathbf{\Phi}_\text{c}^{k}),
\end{equation}
where $\mathcal{L}_{\operatorname{sim}}(\cdot, \cdot)$ is a loss function measuring the similarity between the registered image $\mathbf{W}^{N}$ and the target image $\mathbf{T}$. $\mathcal{R}(\cdot)$ is the regularization term, which determines the smoothness of the deformation, and $\lambda$ is a regularization parameter. 
Here we use the mean square error (MSE) as loss function for 2D face registration task (every source image has its corresponding target image), due to the same pixel intensity distribution between $\mathbf{W}^{N}$ and $\mathbf{T}$. 
For 3D brain MRI registration (all source images share one target image, \ie a template image), we use the negative cross-correlation loss (NCC), which is robust to intensity variations often found across scans and datasets \cite{balakrishnan2018unsupervised}.
We call the registration problem unsupervised if the ground truths of transformation (deformation field $\mathbf{\Phi}$) are not provided in the training set. Otherwise, it is supervised, such as~\cite{yang2017quicksilver, cao2017deformable,sokooti2017nonrigid, krebs2017robust,dai2020dual}. In our proposed method, no transformation ground truth is given during the training.

Deformable image registration is an ill-posed problem, and requires an additional constraint (regularization) to enforce a spatially smooth deformation. Such regularization highly influences the estimated deformation fields as pointed out by previous studies \cite{papiez2014implicit, de2019deep}. For the purpose of both effective and efficient estimation, we use the $\ell_2$-norm of the second-order derivative of $\mathbf{\Phi}_\text{c}^{k}$ as the regularization term:
\begin{equation}
\mathcal{R}(\mathbf{\Phi}_\text{c}^{k})=\sum\|\nabla^{2} \mathbf{\Phi}_\text{c}^{k}\|^{2}.
\end{equation}
More importantly, this enables to better capture the global movement as it puts zero penalty for affine transformations and only non-affine transformation is penalized \cite{rueckert1999nonrigid}.

Benefiting from the differentiability of each component of this design, our model can be cooperatively and progressively optimized across each stage in an end-to-end manner.





%% file: fig_network2.tex
\begin{figure*}[t]
  \centering
  \includegraphics[width=1.0\linewidth]{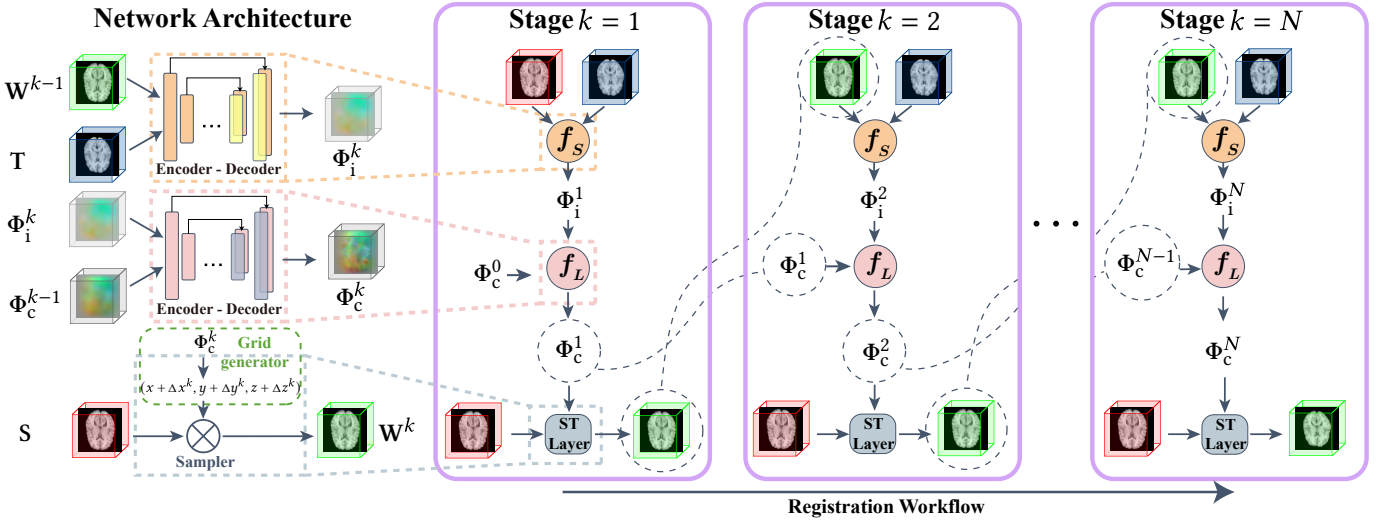}
  \vspace{-10pt}
  \caption{An overview of ABN for multi-stage deformable image registration. In each stage $k$, the \emph{Short-Term Registration Network} $f_{S}$ predicts the current deformation field $\mathbf{\Phi}_\text{i}^{k}$ between the previous warped image $\mathbf{W}^{k-1}$ and target image $\mathbf{T}$; \emph{Long-Term Memory Network} $f_{L}$ fuses current deformation field $\mathbf{\Phi}_\text{i}^{k}$ and previous combined deformation field $\mathbf{\Phi}_\text{c}^{k-1}$ to generate the updated combined deformation field $\mathbf{\Phi}_\text{c}^{k}$; \emph{Spatial Transformation Layer} (ST Layer) performs the nonlinear transformation on the source image $\mathbf{S}$ using $\mathbf{\Phi}_\text{c}^{k}$ to produce the warped image $\mathbf{W}^{k}$. The final warped registered image is $\mathbf{W}^{N}$.}
  \label{fig:network}
  \vspace{-10pt}
\end{figure*}

%% file: 05_experiment_appendix_merged.tex
\section{Experiments}
\label{section:Experiments}

\subsection{Datasets} \label{section:Dataset}
In order to evaluate the performance of our method on image registration tasks, we conduct experiments on three different datasets, one with natural images (2D) and two with medical images (3D).

\noindent\textbullet\ \textit{Flickr-Faces-HQ (FFHQ)} \cite{karras2019style}: 
We used a subset of the FFHQ dataset, which contains 400 human face images, randomly sampled from the whole dataset. 
In order to construct pairs of source and target images, similar to \cite{dai2020dual}, we random transform each face image using a random deformation field. 
We use the deformed image as the target image and the raw image as the source image.  Each image is dealigned using a random deformation field smoothed by a Gaussian filter. We set the standard deviation for Gaussian kernel, $\sigma$ = 18, and a scale factor of displacement, $\alpha$ = 800, to ensure a reasonable deformation.
We resize all images to $64\times64$, and randomly split the image pairs into training and test sets with the ratio of 80$\%$ and 20$\%$, respectively. 

\noindent\textbullet\ \textit{LPBA40} \cite{shattuck2008construction}: 
This dataset consists of 40 T1-weighted 3D brain MRI scans and the corresponding segmentation ground truth of 56 anatomical structures. 
The ground truth is used to evaluate the registration accuracy. 
Same as \cite{balakrishnan2018unsupervised, zhao2019recursive}, we focus on atlas-based registration in this experiment, in which the first scan in the dataset is the target image and the remaining scans need to be aligned with the target image. 
Among the 39 scans,  30 scans are used for training and 9 scans are used for testing. 
All scans are resized to $96\times96\times96$ after cropping. Standard pre-processing steps have been completed in the dataset, including brain extraction and affine spatial normalization.

\noindent\textbullet\ \textit{MindBoggle101} \cite{klein2012101}: 
This dataset contains 101 T1-weighted 3D brain MRI scans and the corresponding annotation of 25 cortical regions, but only 62 scans have their valid segmentation ground truth. 
Following the recent work \cite{liu2019probabilistic}, we focus on atlas-based registration using 41 scans for training and 20 scans for testing. All scans are resized to $96\times96\times96$ after cropping. Same as the LPBA40 dataset, the standard pre-processing steps for the MindBoggle101 dataset have been completed.

\input{tab_methods_v4}
\subsection{Compared Methods}
\label{section:Compared Methods}
We compare our approach to seven state-of-the-art methods, as shown in Table~\ref{tab:methods}.

\noindent\textbullet\ \textit{Affine registration (Affine)} \cite{avants2009advanced}: 
This method breaks the image registration task into a composition of a linear transformation and a translation. We use existing affine implementation in the publicly available software package - ANTsPy~$\footnote{https://github.com/ANTsX/ANTsPy\label{ants}}$ with the default setting.

\noindent\textbullet\ \textit{BSpline transform (BSpline)} \cite{rueckert1999nonrigid}: This method uses control points and spline functions to describe the nonlinear geometric transformation domain. We use existing BSpline implementation in the publicly available software package - SimpleITK (SITK)~$\footnote{https://simpleitk.org/\label{sitk}}$. Each pair of images is optimized with a order of 3, and a gradient tolerance of $1 \times 10^{-10}$ for 200 iterations.

\noindent\textbullet\ \textit{Demons algorithm (Demons)} \cite{thirion1998image}: This method is inspired by the optical flow equations and considers non-rigid image registration as a diffusion process. We use existing Demons implementation in the publicly available software package - SimpleITK (SITK)~$\textsuperscript{\ref {sitk}}$. Each pair of images is optimized with a smooth regularization of 2, and a gradient tolerance of $1 \times 10^{-10}$ for 50 iterations.

\noindent\textbullet\ \textit{Elastic registration (Elastic)} \cite{bajcsy1989multiresolution,avants2009advanced}: This method estimates the elastic geometric deformation by updating transformation parameters iteratively. Elastic is included in the ANTsPy~$\textsuperscript{\ref {ants}}$ software package. We run the elastic registration with the default setting.

\noindent\textbullet\ \textit{Symmetric normalization (SyN)} \cite{avants2008symmetric}: This is a top-performing traditional method for deformable image registration \cite{klein2009evaluation, balakrishnan2018unsupervised}. This method optimizes the space of diffeomorphic maps by maximizing the cross-correlation between images. We run SyN via ANTsPy~$\textsuperscript{\ref {ants}}$ with a default setting.

\noindent\textbullet\ \textit{VoxelMorph (VM)} \cite{balakrishnan2018unsupervised}: This is an unsupervised single-stage registration method, which use one network to predict the deformation between images. For network architectures, we use the latest version, VoxelMorph-2, and configure 10 convolutional layers with 16, 32, 32, 64, 64, 64, 32, 32, 32 and 16 filters. The kernel size of each convolutional layer is $3 \times 3$. The ratio of regularization is set to $\lambda=10$.


\noindent\textbullet\ \textit{Cascaded Registration Networks (CRN)} \cite{zhao2019recursive}:
This is a state-of-the-art method for unsupervised image registration with a multi-stage design. In different stages, the source image is repeatedly deformed to align with a target image. The number of stages is set to 10. In each stage, we configure 10 convolutional layers with 16, 32, 32, 64, 64, 64, 32, 32, 32 and 16 filters. The kernel size of each convolutional layer is $3 \times 3$. The ratio of regularization is set to $\lambda=10$.


\noindent\textbullet\ \textit{Anti-Blur Networks (ABN)}:
This is our proposed model which consists of two sub-networks at each stage, the short-term registration and long-term memory networks. The number of stages is set to 10.
We configure 10 convolutional layers in each sub-network with 16, 32, 32, 64, 64, 64, 32, 32, 32 and 16 filters. The kernel size of each convolutional layer is $3 \times 3$. The ratio of regularization is set to $\lambda=10$.

\noindent\textbullet\ \textit{Anti-Blur Networks-Long (ABN-L)}: ABN-L is a variant of ABN, which only contains a long-term memory network $f_{L}(\cdot, \cdot)$ at each stage. The network handle the registration and combination of deformations simultaneously. The number of stages is set to 10. The long-term memory network has 3 inputs, thus we increase the filters of convolutional layers to 32, 64, 64, 128, 128, 128, 64, 64, 64, and 32. The kernel size of each convolutional layer is $3 \times 3$. The ratio of regularization is set to $\lambda=10$.



\subsection{Evaluation Metrics}
We use two metrics to assess the registered image quality: registration accuracy and image sharpness, which are detailed as follows. 

\subsubsection{Registration Accuracy}
\label{section:Registration accuracy}
The 2D face dataset only contains the pairs of source and target images. To evaluate the registration accuracy, we measure the image similarity by cross-correlation (CC) \cite{rabiner1978digital} and structural similarity (SSIM) \cite{wang2004image} between the final warped image and the target image.

The brain MRI datasets contain the segmentation ground truth of anatomical structures, which is location labels of different tissues in the brain. If the two images are well-aligned, then their anatomical structures should overlap with each other. Thus, it is reasonable to use the structure overlap between the final warped image and the target image to evaluate the registration performance. Similar to \cite{balakrishnan2018unsupervised,zhao2019recursive,de2019deep}, we used Dice score and Jaccard coefficient to measure the overlap between two structures (warped and target).
The final score is the average of the dice score of the each structure.

\subsubsection{Image Sharpness}
We quantify the sharpness of the final warped image by Sum Modulus Difference (SMD) \cite{santos1997evaluation} and Tenengrad \cite{yeo1993autofocusing,tenenbaum1970accommodation}. Those are popular no-reference sharpness evaluation metrics.
SMD evaluated the sharpness of an image by summing the differences between adjacent pixels. If the image is sharp, then we expect the difference is larger than that of a blurred image. 
Tenengrad evaluated the sharpness of an image by extracting its edge information. The image edge is the region where the intensity of the image signal changes rapidly, and the sharper the image, the faster the edge changes. 


\subsection{Experiment Setting.}
We split the datasets into training and test sets as described in the Datasets section. 
For 2D face registration, models are trained with a batch size of 16 for 5$k$ epochs. 
For 3D brain MRI registration, we reduce the batch size to 1 to address GPU memory limitation and models were trained for 1$k$ epochs. 
For both tasks, models are optimized using Adam optimizer~\cite{kingma2014adam} with a learning rate of $1 \times 10^{-4}$. 
The source code is available at \url{https://github.com/anonymous3214/ABN}.

\subsection{Experimental Results}
We evaluate the proposed method in 2D face registration and 3D brain MRI registration tasks. For each task, we quantify the performance of registration accuracy and the image sharpness by their corresponding metrics. Across all of these metrics, we find that ABN achieves higher registration accuracy and image sharpness than the state-of-the-art methods.



\subsubsection{Results on 2D Face Registration}
Table~\ref{tab:2D res} summarizes the results of four methods on 2D face registration task. It can be seen that ABN clearly outperforms all the baselines in terms of both registration accuracy (SSMI and CC) and image sharpness (SMD and Tenengrad). We observed a gain in SSMI of roughly $3.8\%$ and in SMD up to $9.9\%$ compared to the best baseline method CRN.
Notably, the registration accuracy of ABN-L is lower than that of ABN since ABN-L only contains a long-term memory network, which is insufficient for obtaining a high quality registration result.

Figure \ref{fig:face result} shows visual comparisons of 2D face registration results. Upon inspection, we can see that the final warped image of ABN is more similar to the target image than those of VM, CRN and ABN-L. Most notably, in terms of sharpness, ABN-L and ABN clearly outperform CRN. VM does not appear to have a loss in sharpness, but it is significantly worse than other methods in registration accuracy. Furthermore, we also compare the intermediate warped results of our method with other multi-stage method (\ie CRN), as shown in Figure \ref{fig:face stage}. It is clear that both methods allow the warped/registered images to progressively align to the target image with the help of a multi-stage fashion. However, the registration accuracy improvement of CRN is accompanied by a loss of image sharpness, while ABN gives much sharper images than CRN.
\input{fig_res_2d}
\input{tab_2d}

\input{fig_face_stage_v2}
\subsubsection{Results on 3D Brain MRI Registration}
Table \ref{tab:3D res} shows the results of nine methods on two respective datasets: LPBA40 and Mindboggle101.
In this task, Dice score and Jaccard are used to evaluate registration accuracy, while SMD and Tenengrad are used to assess sharpness performance.
The results of each method are shown with its average performance with standard deviation and its ranking among all other methods.
Based on the global competition in both datasets, the average ranking of ABN outperforms that of all baselines. 
Specifically, ABN achieves superior performance against all compared methods on registration accuracy, while achieving competitive results in image sharpness persevering.
We observe that almost all multi-stage methods outperform the single-stage methods on registration accuracy, but only ABN-L and ABN are on par with single-stage methods in terms of image sharpness persevering.
Again, the overall performance of ABN-L was slightly inferior to that of ABN, indicating that the short-term registration network is crucial and beneficial to improve registration accuracy.
\input{tab_3d_v4}

\input{fig_para_3d}
Figure \ref{fig:brain result} shows the visual comparisons of one sample of the 3D brain MRI registration task on the LPBA40 dataset. We can observe that our ABN method is more accurate in aligning the source image to the target image. Notably, although CRN has an impressive performance in registration accuracy, the warped registered image is obviously blurred, which can be verified with the focused example of the anatomical structure. 
Although single-stage methods (Affine, BSpline, Demons, Elastic, SyN, and VM) involve only one interpolation and o dnot suffer from image blurring problem, they are inferior to the multi-stage methods (CRN, ABN-L and ABN) in terms of registration performance.
These results indicate that our proposed method effectively achieves a higher level of registration accuracy while preserving image sharpness.

We also compared the registration speed among the traditional and learning-based methods, as shown in Figure~\ref{fig:time}. The run time is measured on the same machine with a Intel$^{\circledR}$ Xeon$^{\circledR}$ E5-2667 v4 CPU and an NVIDIA Tesla V100 GPU.
To clearly show the time of all baselines in the chart, we omitted method Demons, which takes the longest time for registration: 81.98 seconds on the LPBA40 dataset and 100.95 seconds on the Mindboggle 101 dataset. Upon observation, it is clear that all traditional methods consumes much more time compared to learning-based methods. Specifically, all learning-based methods can achieve real-time registration within one second. Compared with the most accurate non-learning method Elastic, our proposed ABN achieves even higher accuracy and only uses 2.5\% time of Elastic. No GPU implementations of BSpline, Demons, Elastic and SyN have been found in previous works \cite{avants2009advanced,bajcsy1989multiresolution,avants2008symmetric,balakrishnan2018unsupervised,zhao2019recursive}.
\input{fig_res_3d}
\subsubsection{Influence of Parameters.}
We study the performance of all multi-stage methods at different stage number settings using the LPBA40 dataset.
For multi-stage methods, the number of stages corresponds to the depth of the network and the number of warping operations. In other words, more stages mean more refinements of the alignment, which is usually beneficial to the improvement of registration accuracy. However, a larger number of stages also consumes more GPU memory and increase the computation time. To investigate the influence of different stage settings, we compared 10 different versions of the CRN, ABN-L, and ABN, with the number of stages varies from 1 to 10.

As illustrated in Figure \ref{fig:3D Dice} and \ref{fig:3D Jaccard}, both ABN-L and ABN significantly outperform CRN in the registration accuracy as the number of stages increases. In addition, ABN-L and ABN consistently preserve the image sharpness, as observed in Figures \ref{fig:3D SMD} and \ref{fig:3D Tenengrad}. Notably, in this task, the effectiveness of CRN in both registration accuracy and image sharpness decreases significantly when the number of stages increases. This supports our intuition that recursively deforming the image obtained
from the previous stage will result in image blurring and signal attenuation, which causes the registration accuracy to further deteriorate. Thus, our methods ABN-L and ABN outperform CRN in the multi-stage case in both registration accuracy and sharpness preserving.

%% file: tab_methods_v4.tex
\begin{table}[t]
    \centering
    \caption{Summary of compared methods.}
    \label{tab:methods}
    \vspace{-5pt}
    \resizebox{\linewidth}{!}{
    \begin{tabular}{lcccc}
    \toprule
    \textbf{Methods}& \textbf{Anti-blur}& \textbf{Multi-stage}& \textbf{Nonlinear}& \textbf{Deep learning}\\
    \midrule
    Affine { \cite{avants2009advanced}} & \cmark  & \xmark & \xmark & \xmark \\
    BSpline { \cite{rueckert1999nonrigid}} & \cmark  & \xmark & \cmark & \xmark\\
    Demons \cite{thirion1998image}& \cmark  & \xmark & \cmark & \xmark\\
    Elastic \cite{bajcsy1989multiresolution}& \xmark  & \cmark & \cmark & \xmark\\
    SyN \cite{avants2008symmetric} & \xmark  & \cmark & \cmark & \xmark\\
    VM \cite{balakrishnan2018unsupervised}& \cmark  & \xmark & \cmark & \cmark\\
    CRN \cite{zhao2019recursive}& \xmark  & \cmark & \cmark & \cmark\\
    ABN-L (ours)& \cmark  & \cmark & \cmark & \cmark \\
    ABN (ours)& \cmark  & \cmark & \cmark & \cmark\\
    \bottomrule
    \end{tabular}
    }
    \vspace{-5pt}
\end{table}

%% file: fig_res_2d.tex
\begin{figure}[t]
  \centering
  \includegraphics[width=\linewidth]{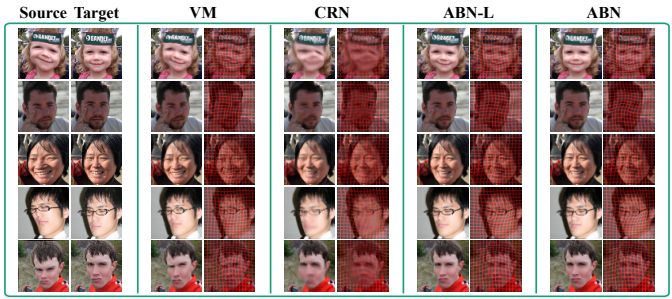}
  \vspace{-5pt}
  \caption{Visual comparisons in 2D face registration task. The first column shows source images, the second column shows target images. For each comparison method, final registered images (left) and deformation grids (right) are shown. }
  \label{fig:face result}
  \vspace{-5pt}
\end{figure}

%% file: tab_2d.tex
\begin{table}[t]
    \centering
    \caption{Results for 2D face registration. The results are reported as performance(mean $\pm$ std). “$\uparrow$” point out “the larger the better”.}
    \label{tab:2D res}
    \vspace{-5pt}
    \resizebox{\linewidth}{!}{
    \begin{tabular}{lcccc}
    \toprule
    \multirow{2}{*}{Methods} & \multicolumn{2}{c}{Registration Accuracy} & \multicolumn{2}{c}{Sharpness} \\
    \cmidrule(lr){2-3}\cmidrule(lr){4-5}
    & SSMI $\uparrow$ & CC $\uparrow$ & SMD $\uparrow$ & Tenengrad $\uparrow$ \\
    \midrule
    VM \cite{balakrishnan2018unsupervised} & 0.653 $\pm$ 0.080 & 0.902 $\pm$ 0.052 & 1.887 $\pm$ 0.398 & 2.382 $\pm$ 0.954\\
    CRN \cite{zhao2019recursive} & 0.921 $\pm$ 0.049 & 0.985 $\pm$ 0.016 & 2.535 $\pm$ 0.537 & 2.518 $\pm$ 0.935\\
    ABN-L (ours) & 0.911 $\pm$ 0.035 & 0.981 $\pm$ 0.012 & 2.719 $\pm$ 0.558 & 2.542 $\pm$ 0.931\\
    \textbf{ABN (ours)} & \textbf{0.956 $\pm$ 0.020} & \textbf{0.992 $\pm$ 0.006} & \textbf{2.786 $\pm$ 0.586} & \textbf{2.565 $\pm$ 0.939}\\
    \bottomrule
    \end{tabular}}
    \vspace{-5pt}
\end{table}

%% file: fig_face_stage_v2.tex
\begin{figure}[t]
  \centering
  \includegraphics[width=\linewidth]{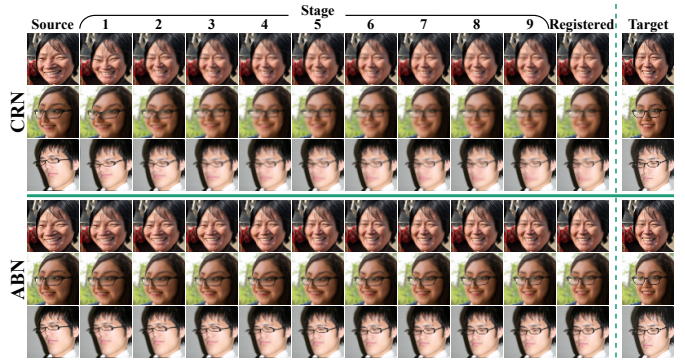}
  \vspace{-10pt}
  \caption{Comparing the intermediate and final results of different multi-stage methods in 2D face image registration. We use the 10-stage CRN and ABN as examples to demonstrate the performance of registration at each stage.}
  \label{fig:face stage}
  \vspace{-10pt}
\end{figure}

%% file: tab_3d_v4.tex
\begin{table*}[t]
    \centering
    \caption{Results for 3D brain MRI registration. The results are reported as performance(mean $\pm$ std {\color{blue}(rank)}). The “Avg. Rank” column shows the average rank of each method across all metric. “$\uparrow$” point out “the larger the better” and “$\downarrow$” point out “the smaller the better”.}
    \label{tab:3D res}
    \resizebox{0.95\linewidth}{!}{
    \begin{tabular}{llcccc c}
    \toprule
    \multirow{2}{*}{Dataset} & \multirow{2}{*}{Methods} & \multicolumn{2}{c}{Registration Accuracy} & \multicolumn{2}{c}{Sharpness} & \multirow{2}{*}{Avg. Rank $\downarrow$}\\
    \cmidrule(lr){3-4}\cmidrule(lr){5-6}
    & & Dice $\uparrow$ & Jaccard $\uparrow$ & SMD $\uparrow$ & Tenengrad $\uparrow$\\
    \midrule
    \multirow{9}{*}{LPBA40} & Affine \cite{avants2009advanced} & 0.631 $\pm$ 0.012 {\color{blue}(9)} & 0.469 $\pm$ 0.016 {\color{blue}(8)} & 0.053 $\pm$ 0.004 {\color{blue}(4)} & 2.846 $\pm$ 0.237 {\color{blue}(3)} & {\color{blue}6.0}\\
    & BSpline \cite{rueckert1999nonrigid} & 0.644 $\pm$ 0.012 {\color{blue}(8)} & 0.480 $\pm$ 0.014 {\color{blue}(7)} & 0.054 $\pm$ 0.004 {\color{blue}(3)} & 2.765 $\pm$ 0.227 {\color{blue}(5)} & {\color{blue}5.8}\\
    & Demons \cite{thirion1998image} & 0.653 $\pm$ 0.064 {\color{blue}(6)} & 0.495 $\pm$ 0.066 {\color{blue}(5)} & 0.054 $\pm$ 0.003 {\color{blue}(2)} & 2.746 $\pm$ 0.355 {\color{blue}(6)} & {\color{blue}4.8}\\
    & Elastic \cite{bajcsy1989multiresolution} & 0.674 $\pm$ 0.012 {\color{blue}(3)} & 0.514 $\pm$ 0.013 {\color{blue}(3)} & 0.051 $\pm$ 0.003 {\color{blue}(5)} & 2.494 $\pm$ 0.190 {\color{blue}(9)} & {\color{blue}5.0}\\
    & SyN \cite{avants2008symmetric} & 0.670 $\pm$ 0.011 {\color{blue}(4)} & 0.510 $\pm$ 0.014 {\color{blue}(4)} & 0.053 $\pm$ 0.004 {\color{blue}(4)} & 2.679 $\pm$ 0.209 {\color{blue}(7)} & {\color{blue}4.8}\\
    & VM \cite{balakrishnan2018unsupervised} & 0.652 $\pm$ 0.017 {\color{blue}(7)} & 0.490 $\pm$ 0.018 {\color{blue}(6)} & 0.055 $\pm$ 0.003 {\color{blue}(1)} & 2.889 $\pm$ 0.194 {\color{blue}(1)} & {\color{blue}3.8}\\
    & CRN \cite{zhao2019recursive} & 0.669 $\pm$ 0.013 {\color{blue}(5)} & 0.510 $\pm$ 0.014 {\color{blue}(4)} & 0.046 $\pm$ 0.002 {\color{blue}(6)} & 2.647 $\pm$ 0.144 {\color{blue}(8)} & {\color{blue}5.8}\\
    & ABN-L (ours) & 0.676 $\pm$ 0.017 {\color{blue}(2)} & 0.518 $\pm$ 0.018 {\color{blue}(2)} & 0.054 $\pm$ 0.003 {\color{blue}(2)} & 2.830 $\pm$ 0.178 {\color{blue}(4)} & {\color{blue}2.5}\\
    & \textbf{ABN (ours)} & 0.684 $\pm$ 0.015 {\color{blue}(1)} & 0.527 $\pm$ 0.017 {\color{blue}(1)} & 0.054 $\pm$ 0.003 {\color{blue}(2)} & 2.847 $\pm$ 0.181 {\color{blue}(2)} & {\color{blue}1.5}\\
    \midrule
    \multirow{9}{*}{Mindboggle101} & Affine \cite{avants2009advanced} & 0.340 $\pm$ 0.017 {\color{blue}(9)} & 0.209 $\pm$ 0.012 {\color{blue}(9)} & 0.077 $\pm$ 0.007 {\color{blue}(3)} & 4.329 $\pm$ 0.775 {\color{blue}(1)} & {\color{blue}5.5}\\
    & BSpline \cite{rueckert1999nonrigid} & 0.349 $\pm$ 0.016 {\color{blue}(8)} & 0.215 $\pm$ 0.012 {\color{blue}(8)} & 0.078 $\pm$ 0.007 {\color{blue}(2)} & 4.247 $\pm$ 0.741 {\color{blue}(2)} & {\color{blue}5.0}\\
    & Demons \cite{thirion1998image} & 0.454 $\pm$ 0.025 {\color{blue}(2)} & 0.300 $\pm$ 0.010 {\color{blue}(2)} & 0.084 $\pm$ 0.008 {\color{blue}(1)} & 4.078 $\pm$ 0.736 {\color{blue}(5)} & {\color{blue}2.5}\\
    & Elastic \cite{bajcsy1989multiresolution} & 0.426 $\pm$ 0.013 {\color{blue}(6)} & 0.275 $\pm$ 0.010 {\color{blue}(6)} & 0.075 $\pm$ 0.007 {\color{blue}(4)} & 3.864 $\pm$ 0.670 {\color{blue}(8)} & {\color{blue}6.0}\\
    & SyN \cite{avants2008symmetric} & 0.399 $\pm$ 0.013 {\color{blue}(7)} & 0.253 $\pm$ 0.010 {\color{blue}(7)} & 0.078 $\pm$ 0.007 {\color{blue}(2)} & 4.025 $\pm$ 0.698 {\color{blue}(6)} & {\color{blue}5.5}\\
    & VM \cite{balakrishnan2018unsupervised} & 0.433 $\pm$ 0.021 {\color{blue}(5)} & 0.281 $\pm$ 0.017 {\color{blue}(5)} & 0.078 $\pm$ 0.007 {\color{blue}(2)} & 4.000 $\pm$ 0.663 {\color{blue}(7)} & {\color{blue}4.8}\\
    & CRN \cite{zhao2019recursive} & 0.440 $\pm$ 0.023 {\color{blue}(4)} & 0.287 $\pm$ 0.018 {\color{blue}(4)} & 0.062 $\pm$ 0.006 {\color{blue}(5)} & 3.339 $\pm$ 0.550 {\color{blue}(9)} & {\color{blue}5.5}\\
    & ABN-L (ours) & 0.444 $\pm$ 0.018 {\color{blue}(3)} & 0.290 $\pm$ 0.015 {\color{blue}(3)} & 0.078 $\pm$ 0.007 {\color{blue}(2)} & 4.084 $\pm$ 0.677 {\color{blue}(4)} & {\color{blue}3.0}\\
    & \textbf{ABN (ours)} & 0.466 $\pm$ 0.030 {\color{blue}(1)} & 0.309 $\pm$ 0.025 {\color{blue}(1)} & 0.078 $\pm$ 0.007 {\color{blue}(2)} & 4.099 $\pm$ 0.696 {\color{blue}(3)} & {\color{blue}1.8}\\
    \bottomrule
    \end{tabular}}
    \vspace{-10pt}
\end{table*}

%% file: fig_para_3d.tex
\begin{figure*}[t]
    \centering
    \subfigure[Accuracy by Dice]{
        \includegraphics[width=0.23\textwidth]{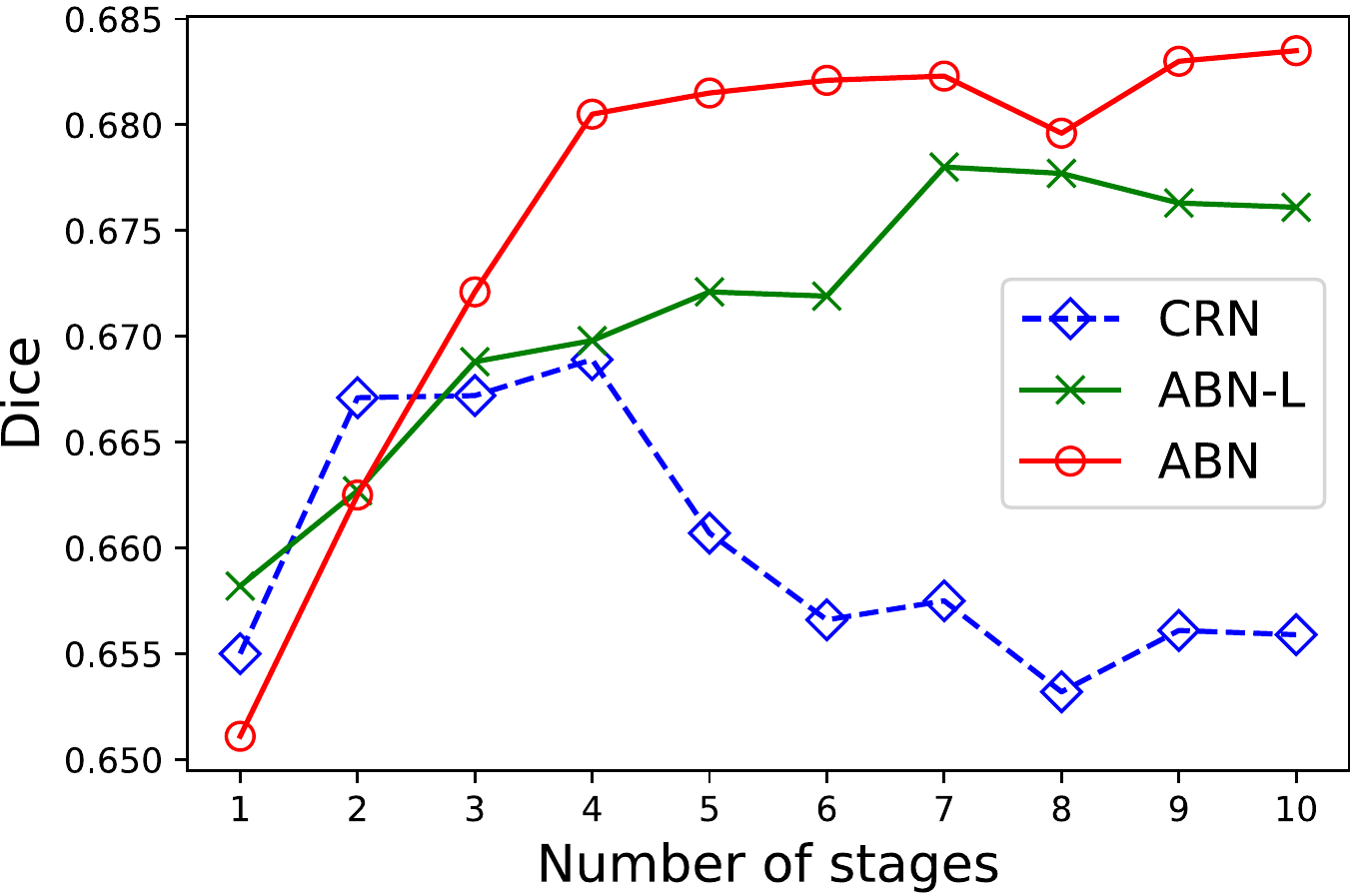}
        \label{fig:3D Dice}
    }
    \subfigure[Accuracy by Jaccard]{
        \includegraphics[width=0.23\textwidth]{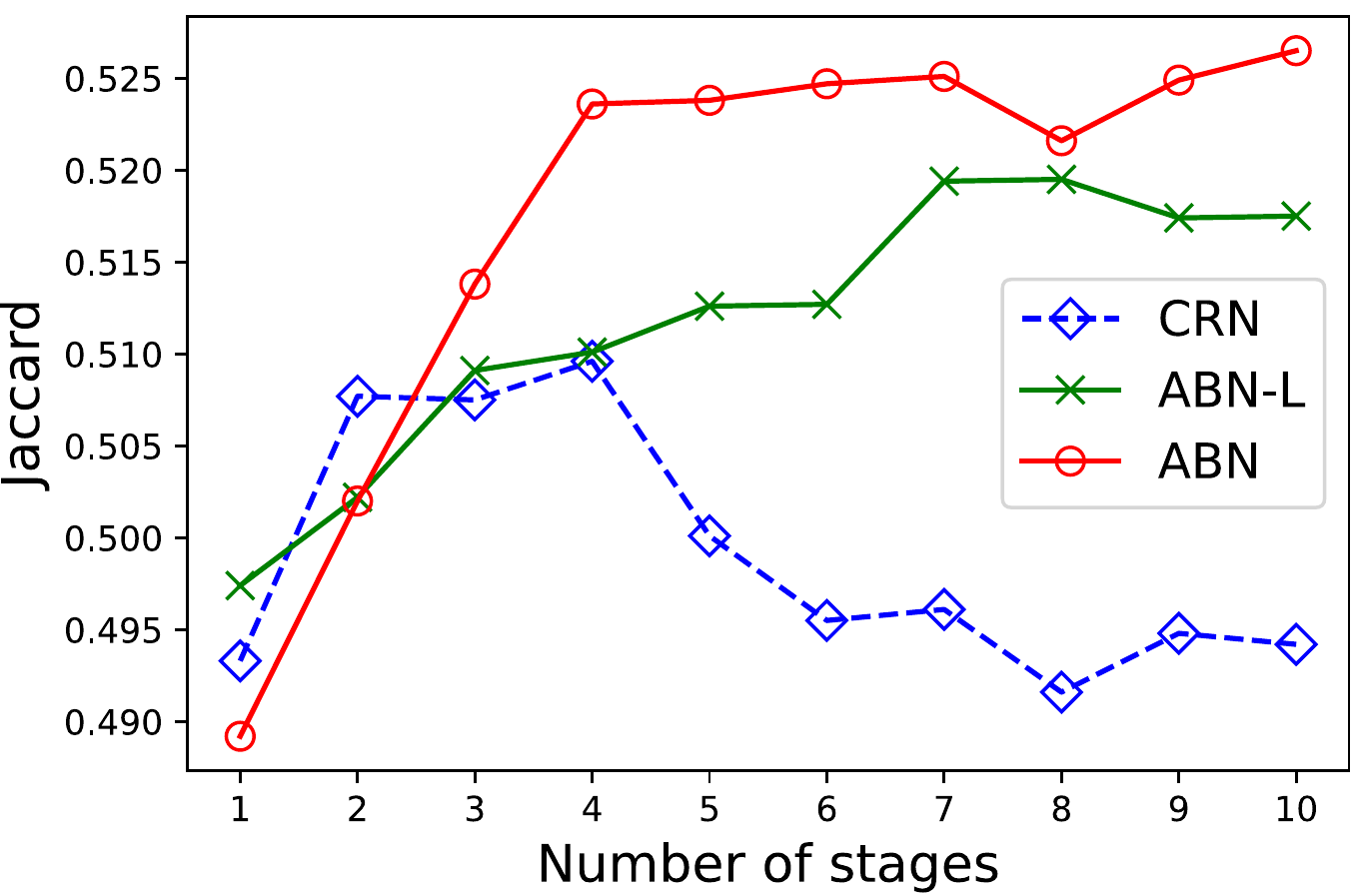}
        \label{fig:3D Jaccard}
    }
    \subfigure[Sharpness by SMD]{
        \includegraphics[width=0.23\textwidth]{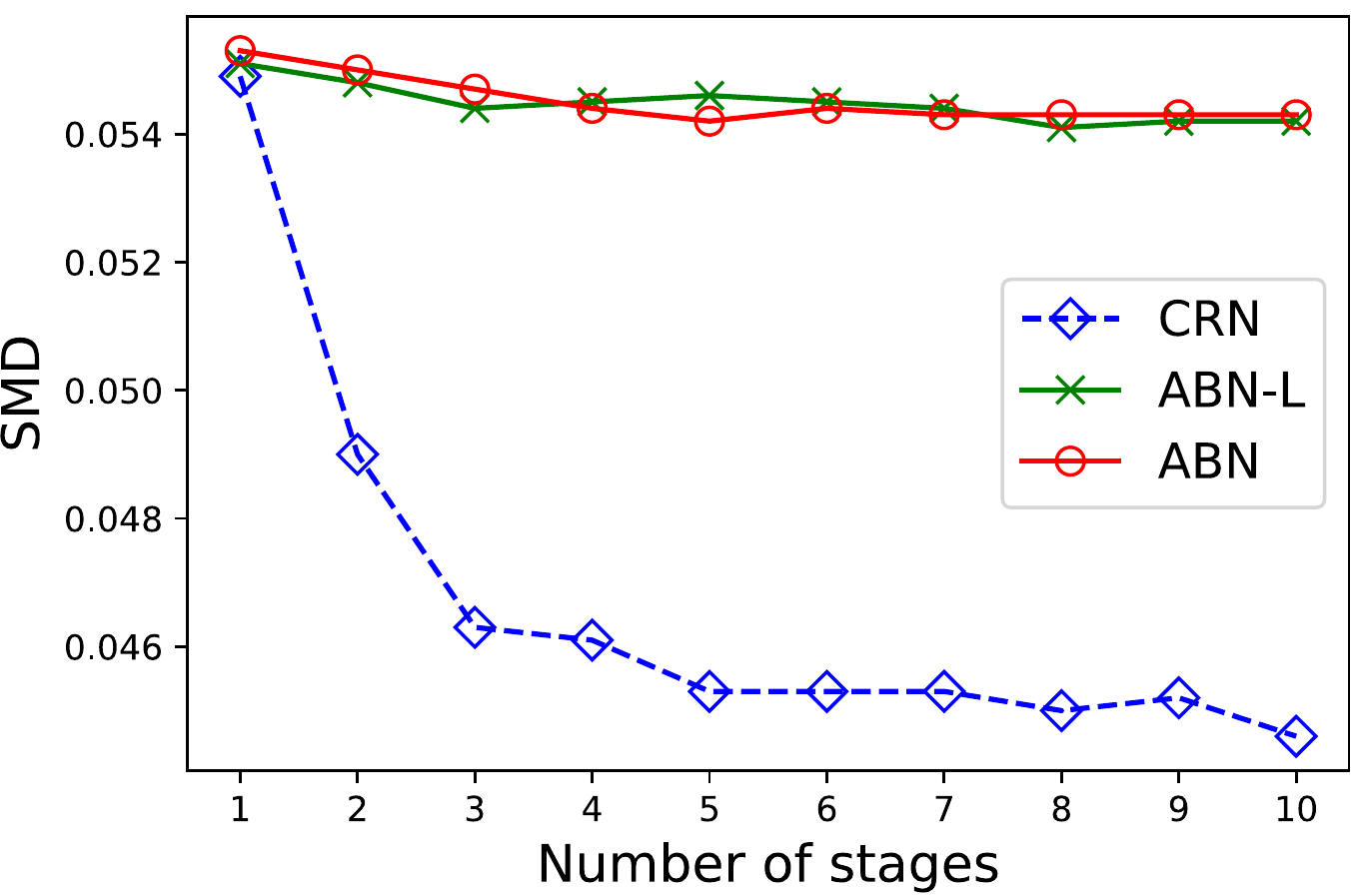}
        \label{fig:3D SMD}
    }
    \subfigure[Sharpness by Tenengrad]{
        \includegraphics[width=0.23\textwidth]{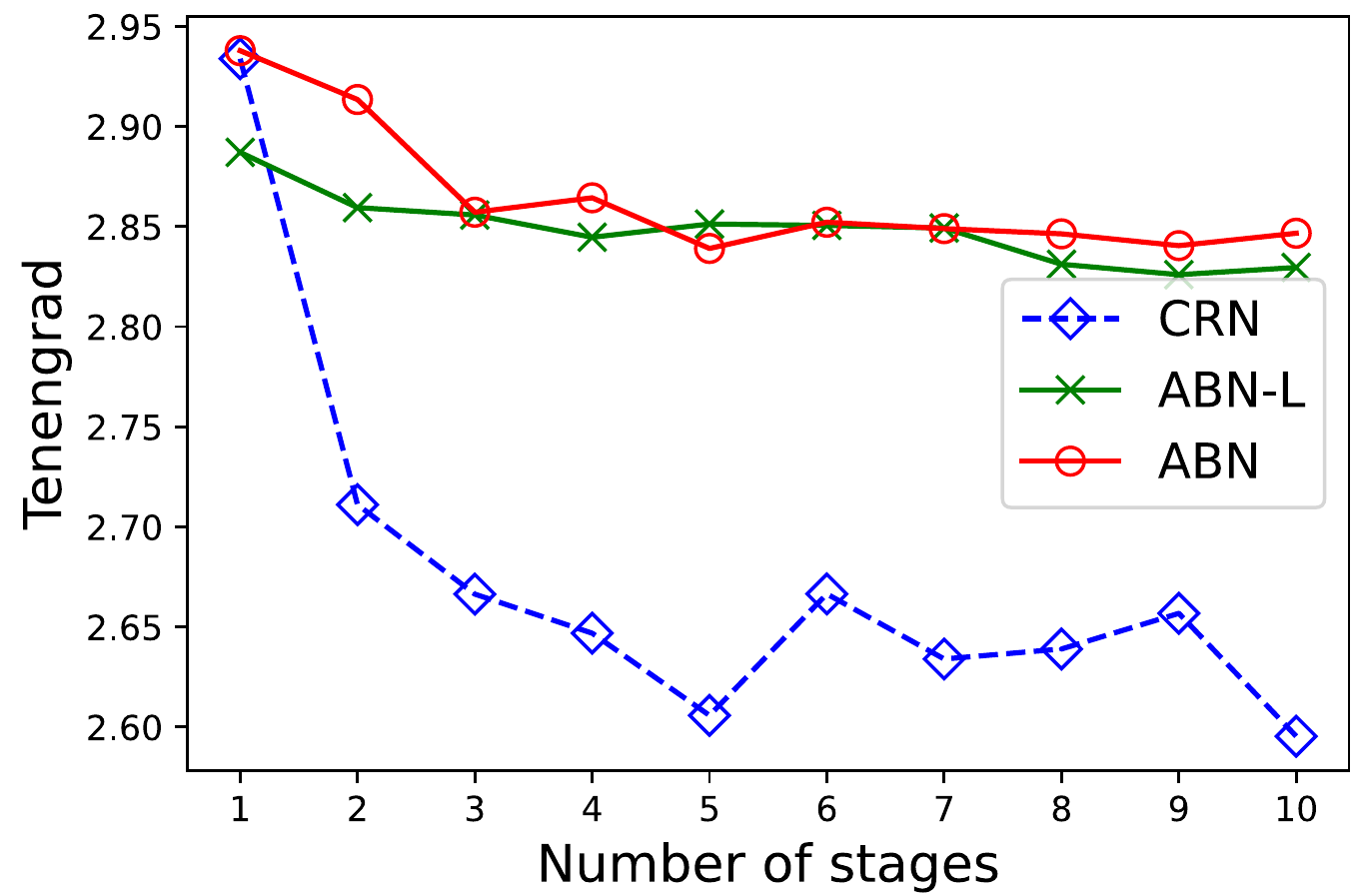}
        \label{fig:3D Tenengrad}
    }
    \vspace{-5pt}
    \caption{Performance of multi-stage methods with different number of stages on 3D brain MRI registration task.}
    \label{fig:3D parameters}
    \vspace{-10pt}
\end{figure*}

%% file: fig_res_3d.tex
\begin{figure*}[t]
  \centering
  \includegraphics[width=\linewidth]{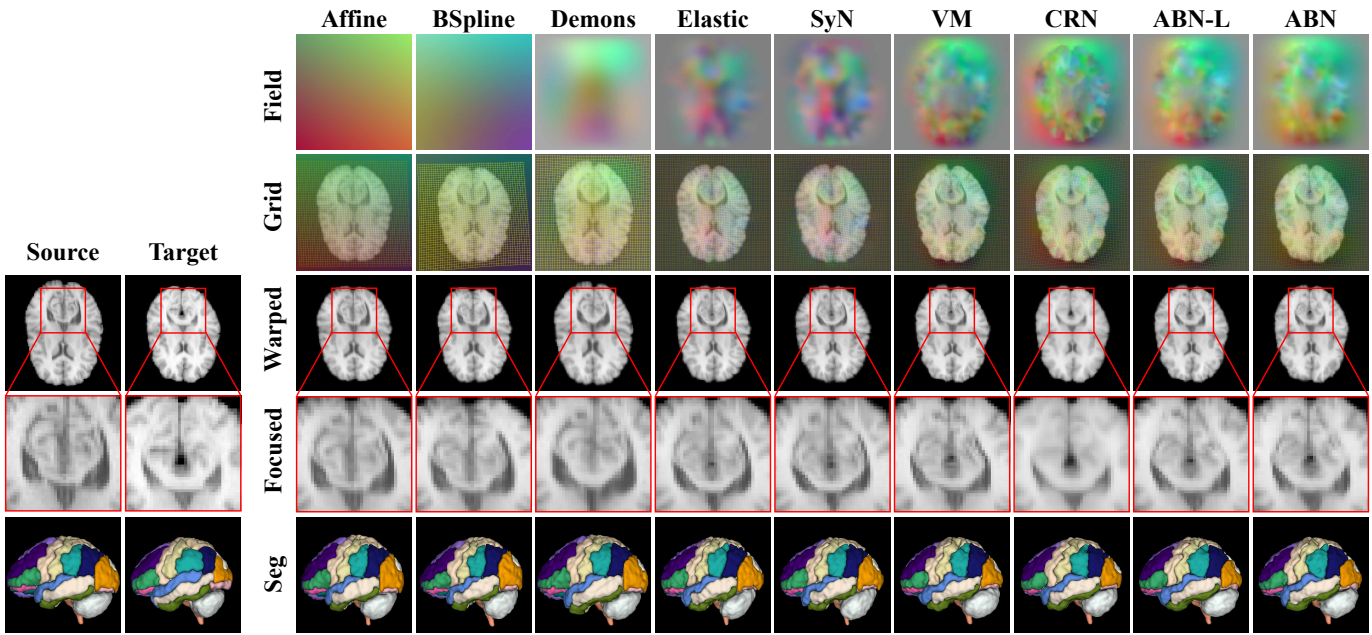}
  \vspace{-5pt}
  \caption{Visual comparisons in 3D brain MRI registration task. The first two columns on the left correspond to the source image and the target image, respectively. From top to bottom, each row shows the predicted deformation field, the deformation grid, the final warped image, a focused example of anatomical structure, and the segmentation label for each comparison method. Deformation fields are displayed by placing the displacements of $x,y,z$ into RGB channels respectively. Segmentation labels are used for the evaluation of registration accuracy.}
  \label{fig:brain result}
  \vspace{-10pt}
\end{figure*}

%% file: 06_related_work.tex
\section{Related Work}
This section briefly reviews the previous works for image registration. 
Traditional image registration methods \cite{avants2009advanced, bajcsy1989multiresolution, avants2008symmetric, jenkinson2001global} optimize similarity between images by iteratively updating transformation parameters. The goal of this optimization is to ensure that the source image and the target image can be aligned to the greatest extent while ensuring the rationality of transformation. To this end, different similarity measures have been proposed for various registration scenarios. For example, Mean Square Error (MSE) on intensity differences is commonly used for image pairs with a similar intensity distribution. However, Normalized Cross-Correlation (NCC) and Mutual Information (MI) usually perform better when it comes to multi-modal registration, \eg brain CT/MRI image registration for tumor localization. Furthermore, it is necessary to smooth the deformation field to avoid folds in the transformation \cite{rueckert1999nonrigid, rohlfing2003volume}. Unfortunately, the optimization of these traditional registration methods is usually performed in a high-dimensional parameter space, which is computationally expensive and time-consuming, limiting their uses in clinical workflows.


To address the potential limitations of traditional image registration, deep learning-based methods \cite{su2022ernet,balakrishnan2018unsupervised,balakrishnan2019voxelmorph,de2017end,zhao2019recursive,de2019deep,yang2017quicksilver,cao2017deformable,sokooti2017nonrigid,krebs2017robust} are being studied more and more extensively in medical image registration. Among them, supervised learning methods depend on the ground truth of transformation mappings. 
In fact, the acquisition of high-quality ground truth in medical practices is often considerately expensive. As a result, the ground truth of supervised learning mainly coming from traditional methods \cite{yang2017quicksilver, cao2017deformable} or synthesis \cite{sokooti2017nonrigid, krebs2017robust,
dai2020dual}. To overcoming this limitation, unsupervised learning approaches have recently gained more attention, which learns image deformation by maximizing the similarity between the warped image and the target image. Benefiting from the success of STN \cite{jaderberg2015spatial}, the gradient can be successfully back-propagated during the training. Several studies registered images by a single-stage design \cite{balakrishnan2018unsupervised,balakrishnan2019voxelmorph,de2017end,li2017non}. 
Nonetheless, it is difficult to achieve desirable registration results with this schema, especially when there are complex and large distortions between images. 
Recently, multi-stage methods \cite{zhao2019recursive, de2019deep} have been proposed to improve the registration accuracy - they decompose a complex transformation step-by-step, recursively warp the source image until that image is well-aligned with the target image. 
However, the above multi-stage registration methods are devoted to improving registration accuracy despite the exposure to information retention failure. 
In this work, we pursue registration accuracy and information retention simultaneously.
\input{fig_time}

%% file: fig_time.tex
\begin{figure}[t]
  \centering
  \includegraphics[width=0.90\linewidth]{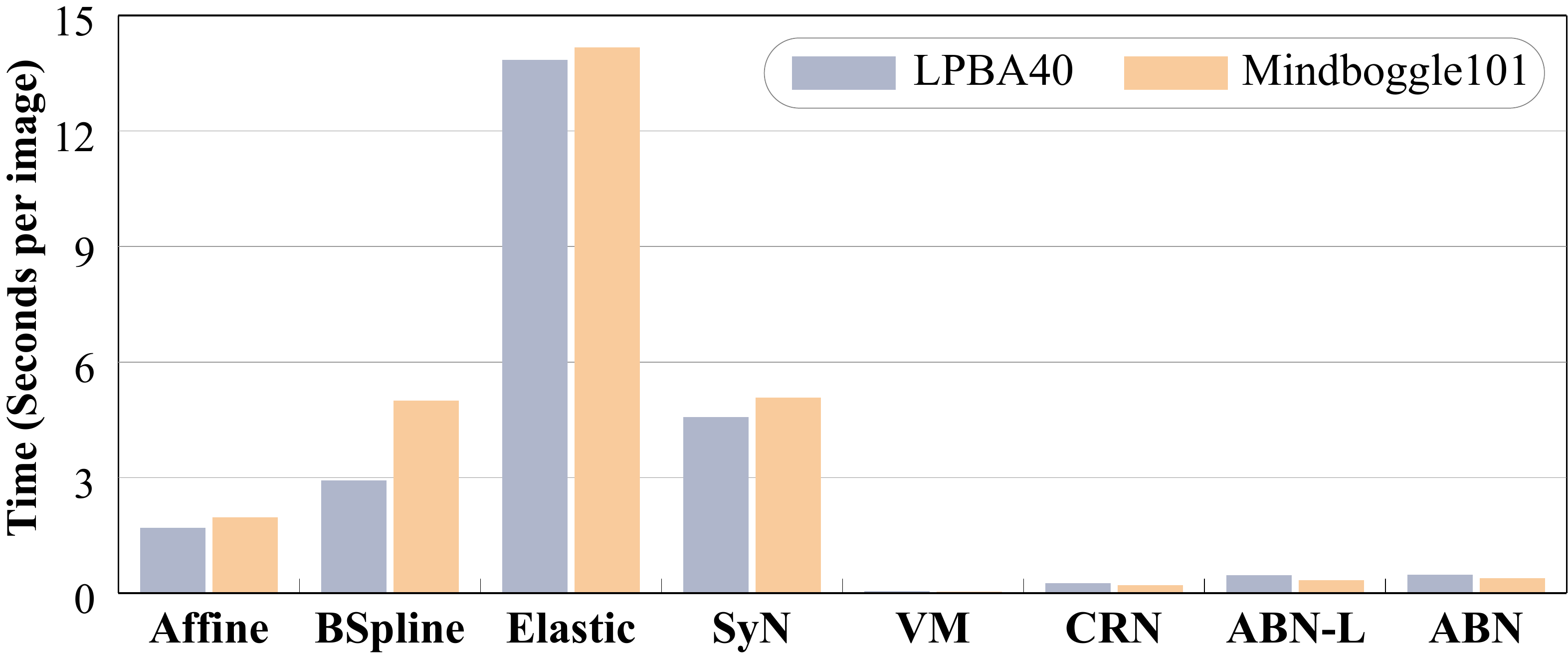}
  \vspace{-5pt}
   \caption{Registration time on 3D brain MRI registration task.}
  \label{fig:time}
  \vspace{-10pt}
\end{figure}

%% file: 07_conclusion.tex
\section{Conclusion}
\label{sec:con}

This paper presents a novel multi-stage neural network method called ABN for anti-blur deformable image registration.
Different from previous works, our proposed method can accurately register images nonlinearly while preserving the image sharpness. 
Specifically, ABN improves the registration accuracy incrementally by a multi-stage design. 
At the same time, a combined deformation of all previous stages is learned simultaneously. 
This deformation is applied directly to the source image to preserve the image sharpness, avoiding information loss caused by multiple interpolations. 
Compared to all other state-of-the-art methods, experimental results demonstrated that ABN consistently generates a comparatively high level of accurate and sharp registered images in both 2D face registration and 3D brain MRI registration tasks.

%% file: 10_ack.tex
\section{Acknowledgments}
\label{sec:ack}
Lifang He was supported by Lehigh's accelerator and core grants.

%% file: fig_res_mind.tex
\begin{figure*}[t]
  \centering
  \includegraphics[width=0.99\linewidth]{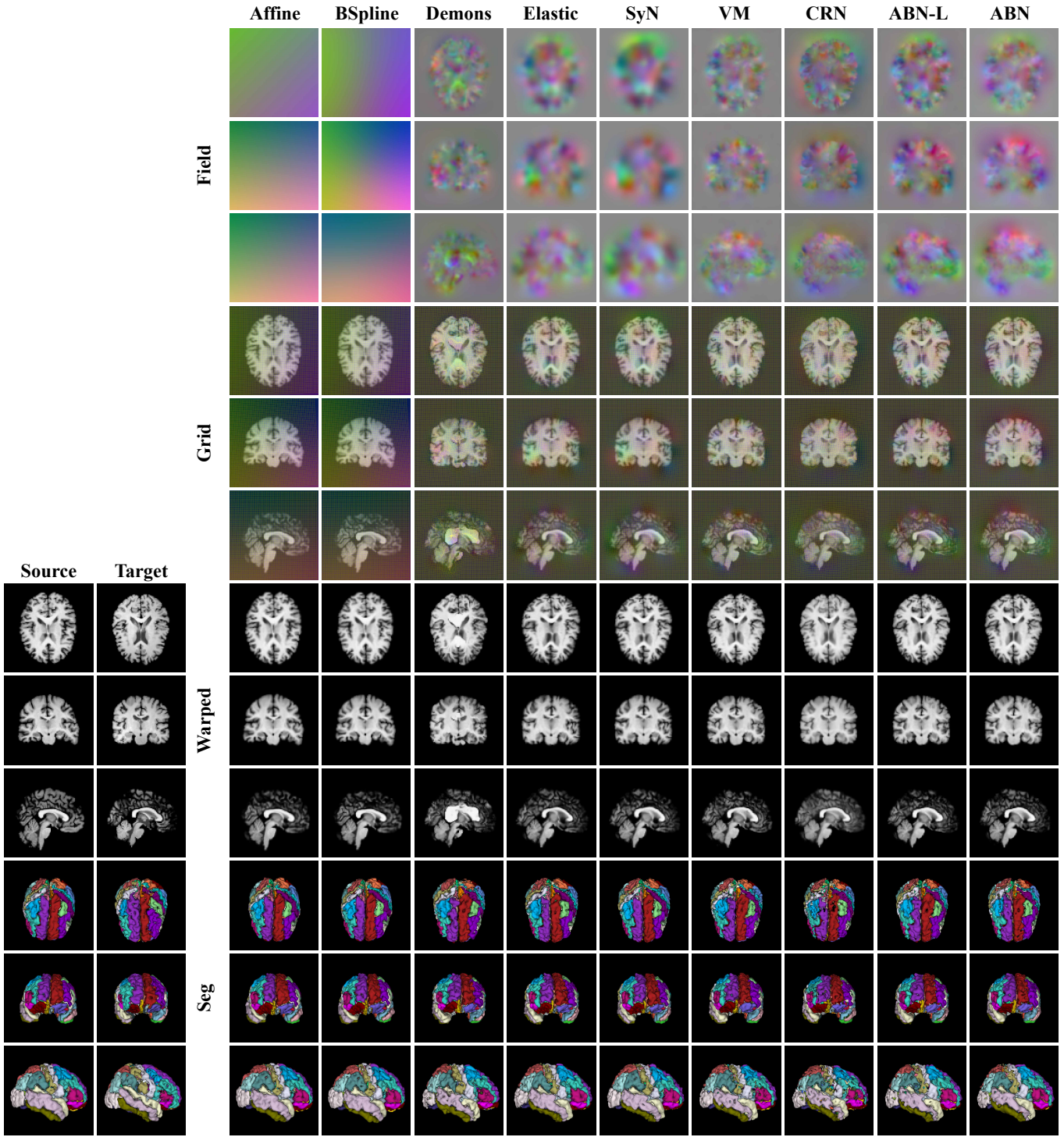}
  \caption{Visual comparison of 3D brain MRI registration on Mindboggle101 dataset. We visualize 3D images by showing the middle plane of each axis. The first two columns on the left correspond to the source image and the target image, respectively. From top to bottom, we provide the predicted deformation field, the deformation grid, the final warped image, and the segmentation label for each comparison method. Deformation fields are displayed by placing the displacements of $x,y,z$ into RGB channels respectively. Segmentation labels are used for the evaluation of registration accuracy.}
  \label{fig:mind}
\end{figure*}